\documentclass[runningheads]{llncs}

\usepackage[T1]{fontenc}
\usepackage{graphicx}
\usepackage{bbding}
\usepackage{subcaption,enumitem}
\usepackage{threeparttable}
\usepackage{multirow}
\usepackage{booktabs}
\usepackage{wrapfig}
\usepackage{amsmath}
\usepackage{lipsum,mwe}
\usepackage{amssymb}
\usepackage[misc]{ifsym}

\begin{document}
\title{Branch Ranking for Efficient Mixed-Integer Programming via Offline Ranking-based Policy Learning}
\titlerunning{Branch Ranking for Efficient Mixed-Integer Programming}
%

\author{Zeren Huang\inst{1} \and
Wenhao Chen\inst{1} \and
Weinan Zhang \inst{1} (\Letter) \and Chuhan Shi\inst{1} \and Furui Liu\inst{2} \and 
Hui-Ling Zhen\inst{2} \and Mingxuan Yuan\inst{2} \and
Jianye Hao\inst{2} \and Yong Yu\inst{1} \and
Jun Wang\inst{2,3}}
\authorrunning{Z. Huang et al.}

\institute{Shanghai Jiao Tong University \and Huawei Noah's Ark lab \and University College London\\
\email{\{sjtu\_hzr, wnzhang\}@sjtu.edu.cn,\{liufurui2, zhenhuiling2\}@huawei.com}} 

\maketitle              

\begin{abstract}
Deriving a good variable selection strategy in branch-and-bound is essential for the efficiency of modern mixed-integer programming (MIP) solvers. 
With MIP branching data collected during the previous solution process, learning to branch methods have recently become superior over heuristics. As branch-and-bound is naturally a sequential decision making task, one should learn to optimize the utility of the whole MIP solving process instead of being myopic on each step.
In this work, we formulate learning to branch as an offline reinforcement learning (RL) problem, and propose a long-sighted hybrid search scheme to construct the offline MIP dataset, which values the long-term utilities of branching decisions. During the policy training phase, we deploy a ranking-based reward assignment scheme to distinguish the promising samples from the long-term or short-term view, and train the branching model named \textsc{Branch Ranking} via offline policy learning. Experiments on synthetic MIP benchmarks and real-world tasks demonstrate that \textsc{Branch Ranking} is more efficient and robust, and can better generalize to large scales of MIP instances compared to the widely used heuristics and state-of-the-art learning-based branching models.

\end{abstract}

\section{Introduction}
Mixed-integer programming (MIP) has a wide range of real-world applications such as scheduling, manufacturing and routing \cite{zhu2000minimizing,maravs2013routing}. A universally applicable method for solving MIPs is branch-and-bound (B\&B) \cite{morrison2016branch}, which performs decomposition of the solution set iteratively, and thus building a search tree with nodes corresponding to MIP problems. 

Node selection and variable selection are two important sequential decisions to be made at each iteration of the B\&B algorithm. The node selection strategy decides which node to process next, while the variable selection strategy decides which fractional variable to branch on at the current node. In this work, we focus on variable selection, which can dramatically influence the performance of the B\&B algorithm, as discussed in \cite{morrison2016branch}.
 
The variable selection strategies in modern MIP solvers are generally based on manually designed heuristics, which are strongly dependent on the problem property. Therefore, when the structure or the scale of the problem changes, an adjustment of the heuristics is often required, which is time-consuming and labor-intensive \cite{fischetti2010heuristics}. 
To tackle the above issues, machine learning naturally becomes a candidate for constructing a more efficient and generalizable variable selection policy \cite{bengio2021machine}. The similarity among many previous learning-based methods lies in the way to train the branching policy \cite{khalil2016learning,alvarez2017machine,gasse2019exact,gupta2020hybrid,nair2020solving}. Specifically, the policy models are usually trained via imitation learning to mimic an effective but slow heuristic, i.e., \emph{strong branching} (SB), which performs a one-step branching simulation for each candidate variable and greedily selects one with the highest one-step utility, the SB score, which is related to the dual bound improvements. As B\&B is naturally a sequential decision making task, such a heuristic is considered to be myopic, that is, it only focuses on the short-term utility while neglecting the long-term impact on the bottom levels of the B\&B tree. Under some circumstances, the optimal short-term branching decision can lead to poor long-term utility. 

To address the above issues, in this paper, we formulate variable selection in B\&B as an offline reinforcement learning (RL) problem, and propose a top-down hybrid search scheme to construct an offline dataset which involves branching information of both long-term and short-term decisions. Using the offline dataset, we deploy a ranking-based reward assignment scheme to distinguish the promising samples from other samples. Finally, the variable selection policy, which is named \textsc{Branch Ranking} (BR), is trained via ranking-based policy learning method, which is derived equivalent as maximizing the log-likelihood of samples with the corresponding rewards as the weights. 

We conduct extensive experiments on four classes of NP-hard MIP benchmarks and deploy our proposed policy in the real-world supply and demand simulation tasks, and the results demonstrate that \textsc{Branch Ranking} is superior over widely used heuristics as well as offline state-of-the-art learning-based branching models, regarding the solution time and the number of nodes processed in the B\&B tree. Furthermore, the evaluation results on different scales of MIPs show that \textsc{Branch Ranking} also has better generalization ability over larger scales of problems compared to other learning-based policies. 

\textsc{Branch Ranking} has been deployed in real-world \emph{demand and supply simution} tasks of Huawei, a global commercial technology enterprise, where the experiments demonstrate that \textsc{Branch Ranking} is applicable in decision problems encountered in practice.

In summary, the main technical contributions of this work are threefold:
\begin{itemize}
    \item We formulate the variable selection in B\&B as an offline RL task, and design a novel search scheme to construct the offline dataset which involves both long-term and short-term branching information. 
    \item Based on the constructed offline dataset, we propose a ranking-based reward assignment scheme to distinguish the promising samples, and train the variable selection policy named \textsc{Branch Ranking}.
    \item Extensive experiments demonstrate that \textsc{Branch Ranking} can make better improvements to the optimization algorithm compared to other state-of-the-art heuristics and learning-based branching polices. \textsc{Branch Ranking} also shows better generalization ability to MIP problems with different scales.
\end{itemize}

\section{Related Work}
The branching decision is one of the most important decisions to be made in the MIP solvers, which can significantly influence the efficiency of the optimization algorithm. The core of branching for mixed-integer programming (MIP) is the evaluation of the branching scores and cost via tuning hyper-parameters. It is not surprising that the most of previous branching methods are devoted to imitate strong branching (SB), because with one step forward, SB can often effectively reduce the number of search tree nodes.

Khalil et al.~\cite{khalil2016learning} extract 72 branching features artificially including static and dynamic ones, and mimicked the strong branching strategy via a pair-wise variable ranking formulation. Similarly, our work also adopts a ranking-based scheme, however, we apply it in reward assignment which assigns higher implicit rewards to the top-ranking promising branching samples from the long-term or short-term view, and the assigned reward can be regarded the sample weight in the learning objective. More recently, Gasse et al.~\cite{gasse2019exact} train their policy model via imitation learning from the strong branching expert rules. The inspiring innovation is a novel graph convolutional network (GCN) integrated into Markov decision process (MDP) for variable selection which is beneficial to generalization and branching cost. Based on a similar policy architecture and learning method, Nair et al.~\cite{nair2020solving} propose neural branching, which enables the expert policy to scale to large MIP instances through hardware acceleration. To attain a computationally inexpensive model compared to GCN \cite{gasse2019exact}, Gupta et al.~\cite{gupta2020hybrid} propose a hybrid architecture which combines the expressive GCN and computationally inexpensive multi-layer perceptrons (MLP) for efficient branching with limited computing power. With the aim at learning a branching policy that generalizes across heterogeneous MIPs, Zarpellon et al.~\cite{zarpellon2021parameterizing} incorporate an explicit parameterization of the state of
the search tree to modulate the branching decision. From another perspective, Balcan et al.~\cite{balcan2018learning} propose to learn a weighted version of several existing variable scoring rules to improve the branching decision. Different from the previous works, Sun et al.~\cite{sun2020improving} employ reinforcement learning with a novelty based evolutionary strategy for learning a branching policy. Another reinforcement learning approach to tackle the branching decision problem comes from ~\cite{etheve2020reinforcement}, which uses approximate Q-learning and the subtree size as value function. Note that the main algorithms of the above two reinforcement learning methods are on-policy, which evaluates and improves the same policy for making decisions, thus cannot leverage the abundant pre-collected branching data to train the model.

\section{Background and Preliminaries}
In this section, we first introduce the background of mixed-integer programming and the branch-and-bound algorithm. Then, we introduce several typical and effective variable selection heuristics for branching.

\subsection{Mixed-Integer Programming Problem}
The general combinatorial optimization task is usually formulated as a mixed-integer programming (MIP) problem, which can be written as the following form:
\begin{equation}
\label{mip_formulation}
	\underset{\mathbf{x}}{\arg \min}\left\{\mathbf{z}^{\top} \mathbf{x} \mid \mathbf{A} \mathbf{x} \leq \mathbf{b},  \mathbf{x} \in \mathbb{Z}^{p} \times \mathbb{R}^{n-p}\right\},
\end{equation} 
where $\mathbf{x}$ is the vector of $n$ decision variables, $\mathbf{x} \in \mathbb{Z}^{p} \times \mathbb{R}^{n-p}$ means $p$ out of $n$ variables have integer constraints, $\mathbf{z} \in \mathbb{R}^n$ is the objective coefficient vector, $\mathbf{b} \in \mathbb{R}^{m}$ is the right-hand side vector, $\mathbf{A} \in \mathbb{R}^{m \times n}$ is the constraint matrix.

\subsection{Branch-and-Bound}
\label{sec-bb}
B\&B is a classic approach for solving MIPs, which adopts a search tree consisting of nodes and branches to partition the total set of feasible solutions into smaller subsets. 

The procedure of B\&B can be described as follows: denote the optimal solution to the LP relaxation of Eq.~(\ref{mip_formulation}) as $x^*$, if it happens to satisfy the integrality requirements, then it is also the solution to Eq.~(\ref{mip_formulation}); else some component of $x^*$ is not integer (while restricted to be integer), then one can select a fractional variable $x_i$ and decompose the LP relaxation into two sub-problems by adding rounding bounds $x_i \geq \lceil x_i^* \rceil$ and $x_i \leq \lfloor x_i^* \rfloor$, respectively. By recursively performing the above binary decomposition, B\&B naturally builds a search tree, within which each node corresponds to a MIP.

During the search process, the best feasible solution of the MIP provides an upper bound (or primal bound) for the optimal objective value; and solving the LP relaxation of the MIP provides the lower bound (or dual bound). The B\&B algorithm terminates when the gap between the upper bound and the lower bound reduces to some tolerance threshold.

\subsection{Variable Selection Heuristics for Branching}
In the B\&B framework for solving MIPs, for a given node, variable selection refers to the decision to select an integer variable to branch on. Designing a good variable selection strategy is essential for the efficiency of solving MIPs, which can lead to a much smaller search tree, that is, the number of processed nodes is significantly reduced and thus the whole solution process speeds up.

In modern MIP solvers, the variable selection module for branching is regarded as a core component of the optimization algorithm, and is generally based on manually designed heuristics. One of the classic heuristics is \emph{strong branching}, which involves computing the dual bound improvements for each candidate variable by solving two resulting LP relaxations after temporarily adding bounds. Strong branching can often yield the smallest search trees while bringing much more computational costs, therefore, it is impractical to apply strong branching at each node. 

Another more efficient heuristic, \emph{pseudo cost branching}, is designed to imitate strong branching, which estimates the dual bound improvements of candidate variables based on historical information gathered in the tree. The combination of the above two heuristics is called \emph{hybrid branching}, which employs strong branching at the beginning of the algorithm, and performs pseudo cost branching when more branching history information is available. 

The current state-of-the-art variable selection heuristic is \emph{reliability branching}, which is a refinement of pseudo cost branching, and is deployed as the default branching strategy by many modern MIP solvers (e.g., CPLEX \cite{cplex2009v12}, SCIP \cite{GleixnerEtal2018ZR}, etc). During the solution process, reliability branching applies strong branching to those variables whose pseudo costs are uninitialized or unreliable. The pseudo costs are considered to be unreliable if there is not enough historical branching data aggregated.

\section{Methodology}
In this section, we first formulate learning variable selection strategy for branching as an offline RL problem, and introduce our method to derive a more efficient and robust variable selection policy named \textsc{Branch Ranking}.

\subsection{Offline RL Formulation}
The variable selection problem in branch-and-bound can be formulated as a sequential decision-making process with tuple $(S, A, \Pi, g)$, where $S$ is the state space, $A$ is the action space, $\Pi$ is the policy space and the function $g(s,a)$ is the dynamics function of the environment. At time step $t \geq 0$, a policy $\pi \in \Pi$ maps the environment state $s_t \in S$ to an action $a_t \in A$: $\pi(s_t) = a_t$, then the next state is $s_{t+1} = g(s_t, a_t)$. In the following, we clarify the state space $S$, the action space $A$, the transition function $g(s,a)$ and the roll-out trajectory $\tau$ in the context of branch-and-bound.

\paragraph{State Space $S$.} The representation of $S$ consists of the whole search tree with previous branching decisions, the LP solution of each node, the processing leaf node and also other statistics stored during the solution process. In practice, we encode the state using GCN as \cite{gasse2019exact}.

\paragraph{Action Space $A$.} At the $t^{\text{th}}$ iteration, the action space $A$ contains all the candidate variables $X^{t}_{cand} = \{x^t_1, x^t_2, \ldots, x^t_d\}$ to branch on at the currently processing node.

\paragraph{Transition Function $g(s,a)$.} After a policy $\pi$ selects an action $a_t = x \in X^{t}_{cand}$ in state $s_t$, the search tree is expanded and the LP relaxations of the two sub-nodes are solved. Then, the tree is pruned if possible, and eventually the next leaf node to process is selected. The environment then proceeds to the next state $s_{t+1}$.

\paragraph{Trajectory $\tau$} A roll-out trajectory $\tau$ comprises a sequence of states and actions: $\{s_0, a_0, s_1, a_1, \ldots, s_T, a_T\}$. We define the return $R_\tau$ for the trajectory $\tau$ as the negative value of the number of visited (processed) nodes within it. Intuitively, each trajectory corresponds to a B\&B tree. Note that there is no explicit reward defined for each state-action pair.

Based on the above MDP formulation, online RL algorithms usually suffer from low sample efficiency, which makes the training prohibitively long and leaves the pre-collected branching data of no use. To derive a more efficient scheme, we model the learning problem in the offline RL settings, where a batch of $m$ roll-out trajectories $D = \{(\tau_i, R_{\tau_i}), i=1,2,\ldots, m\}$ is collected using some policy $\pi_D$. Using this dataset (without interaction with the environment), our objective is to construct a policy $\pi(a|s)$ which incurs the highest trajectory return when it is actually applied in the MDP.

\subsection{Collecting Offline Data through Top-Down Hybrid Search}
\begin{figure}[htb]
     \centering
     \includegraphics[width=\textwidth]{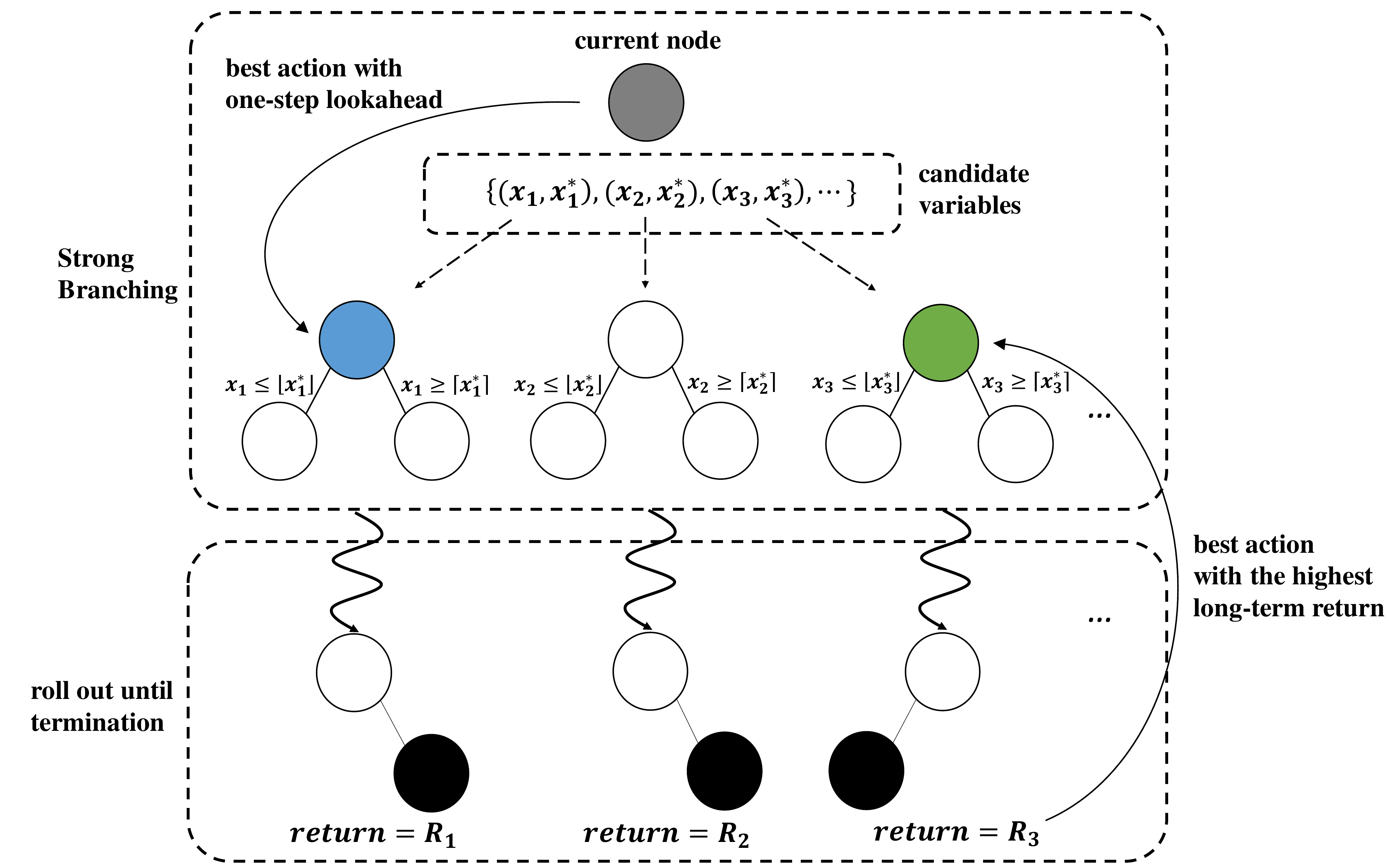}
     \caption{Hybrid offline data collection}
     \label{fig-hybrid}
 \end{figure}

To construct the offline dataset, a commonly-used roll-out policy is strong branching, which is time-consuming while often resulting in a small search tree. Under this case, the roll-out policy can be regarded as the expert policy, which directly provides the supervised data. As shown in the top dotted box in Figure \ref{fig-hybrid}, the blue node represents the action selected by strong branching, which has the largest dual bound improvement after a one-step branching simulation. Strong branching can be viewed as a greedy method for variable selection, therefore, we consider such a strategy to be myopic, that is, it only focuses on the one-step return, which can lead to poor long-term outcome in some cases. As discussed in \cite{glankwamdee2006lookahead}, a deeper lookahead can often be useful for making better branching decisions at the top levels of the search tree.

To derive a more farsighted and robust variable selection strategy for branching, we introduce a top-down hybrid search scheme to collect offline branching data which contains more information about the long-term impact of the branching decision. Figure \ref{fig-hybrid} presents an overview of our proposed hybrid search scheme, which is a combination of short-term and long-term search. The basic procedure is as follows:

\begin{itemize}
    \item At the $t^{\text{th}}$ iteration, for the currently processing node, the candidate variable set is $X^{t}_{cand} = \{x^t_1, x^t_2, \ldots, x^t_d\}$. First, we randomly sample the variables in $X^t_{cand}$ for $k$ times, which returns a subset of $k$ variables $X^{t}_{exp} = \{x^{t}_{e_1}, x^{t}_{e_{2}}, \ldots, x^{t}_{e_k}\} $. 
    \item For each variable $x \in X^{t}_{exp}$, we perform a one-step branching simulation, and obtain $k$ simulation trees with different branches at the current node. Denote the set of $k$ simulation trees as $T^{t}_{exp} = \{T^t_{e_1}, T^t_{e_2}, \ldots, T^t_{e_k}\}$.
    \item As it shown in the lower dotted box in Figure \ref{fig-hybrid}, for each simulation tree $T^t_{e_i} \in T^{t}_{exp}$, we continue the branching simulation, that is, to roll out the trajectory using some myopic policy (e.g. strong branching) until the termination node (black nodes), which returns a trajectory return $R_{e_i}^t$. Then, we obtain a set consisting of pairs of the top-level exploring action and the long-term trajectory return: $B^t = \{(x_{e_i}^t, R_{e_i}^t), i=1,2,\ldots, k\}$. 
    \item Back up at the node processing at the $t^{\text{th}}$ iteration, we select the action $x \in X^t_{exp}$ with the highest long-term return in $B^t$ (the green node in Figure \ref{fig-hybrid}), and execute the corresponding branching in the real environment. The search tree then picks the next node to process using the node selection policy incorporated in the solver, and then the environment transits to the next state. The above process continues until the problem instance is solved.
\end{itemize}

Note that though collection branching data leads to multiple rounds of branching simulation, the whole process is conducted in an offline way, and thus the incurred cost is acceptable for practical use. Denote the collected offline dataset using the above hybrid search scheme as $D_{hyb} = \big \{\{(s^L_i, a^L_i, R^L_i)\}_{i=1}^{M} \cup \{(s^{SB}_i, a^{SB}_i)\}_{i=1}^{N} \big \}$, in which $D_{L} = \{(s^L_i, a^L_i, R^L_i)\}_{i=1}^{M}$ is collected by the top-down long-term search as mentioned above, and $D_{SB} = \{(s^{SB}_i, a^{SB}_i)\}_{i=1}^{N}$ is collected by strong branching during the simulation process.

\subsection{Ranking-based Policy Learning}
Given the offline dataset $D_{hyb} = \{D_L \cup D_{SB}\}$ collected by our proposed hybrid search scheme, our goal is to derive a efficient and robust variable selection policy $\pi_{\theta}$ ($\theta$ is the policy parameter) which can yield a smaller search tree, and thus leading to faster performance.  

Assume that the optimal policy for branching is $\pi^*$, and if $\pi^*$ is available, the policy $\pi$ can be trained via minimizing the KL divergence of $\pi^*$ and $\pi$, which is also equivalent to maximizing the log-likelihood:
\begin{equation}
    \min _{\theta} KL\left[\pi^* \| \pi_{\theta}\right] = \max_{\theta} \mathbb{E}_{(s,a) \sim D_{exp}} \left[\pi^* (a | s) \log \pi_{\theta} (a | s) \right],
\end{equation}
where $D_{exp}$ is the roll-out dataset collected by $\pi^*$. We regard $\pi^*(a | s)$ as the implicit reward $\hat{r}(s,a)$ of the state-action pair $(s,a)$, which measures the implicit value of taking action $a$ in state $s$ in the branch-and-bound tree. Assume that the state-action pairs are sampled from the offline dataset, the learning objective can be approximated in the following form:
\begin{equation}
\label{obj}
    \max_{\theta} \mathbb{E}_{(s,a) \sim D_{hyb}}\left[\hat{r}(s,a) \log \pi_{\theta} (a | s)\right].
\end{equation}

Using the offline dataset $D_{hyb} = \{D_L \cup D_{SB}\}$, we propose a simple while effective ranking-based implicit reward assignment scheme. To derive a policy which can balance the short-term and long-term benefits, we assign higher implicit rewards to the promising state-action pairs which lead to the top-ranked long-term or short-term return. We first give a ranking-based definition of long-term and short-term promising state-action pairs in Definition \ref{ltp} and \ref{stp}, respectively.

\begin{definition}[Long-Term Promising]
\label{ltp}
    For a state-action pair $(s,a)$, its corresponding trajectory return $R^L(s,a)$ is the return of the trajectory starting from state $s$ and selecting the action $a$. A state-action pair $(s,a)$ is defined to be long-term promising if $R^{L}(s,a)$ ranks in the top $p\%$ trajectories starting from state $s$ and with the highest trajectory returns. Note that $p$ is a tunable hyper-parameter.
\end{definition}

\begin{definition}[Short-Term Promising]
\label{stp}
    A state-action pair $(s,a)$ is short-term promising if the one-step look-ahead return of the action $a$ ranks the highest in state $s$ in terms of strong branching score, which is related to the dual bound improvements after a one-step simulation.
\end{definition}
Based on the above ranking-based definitions, for a state-action pair $(s,a)$, its implicit reward $\hat{r}(s,a)$ is defined as
\begin{equation}
    	 \hat{r}(s,a) = \left\{
	\begin{array}{lcl}
    		1, & \quad \mbox{$(s,a)$ is long-term or short-term promising}\\
		0, & \quad \mbox{otherwise.}
	\end{array}
	\right.
\end{equation}
After the reward assignment, the final constructed training dataset is a combination of long-term and short-term promising samples. We denote the proportion of short-term promising samples as $h$, which is a tunable hyper-parameter.

\begin{figure}[htb]
     \centering
     \includegraphics[width=\textwidth]{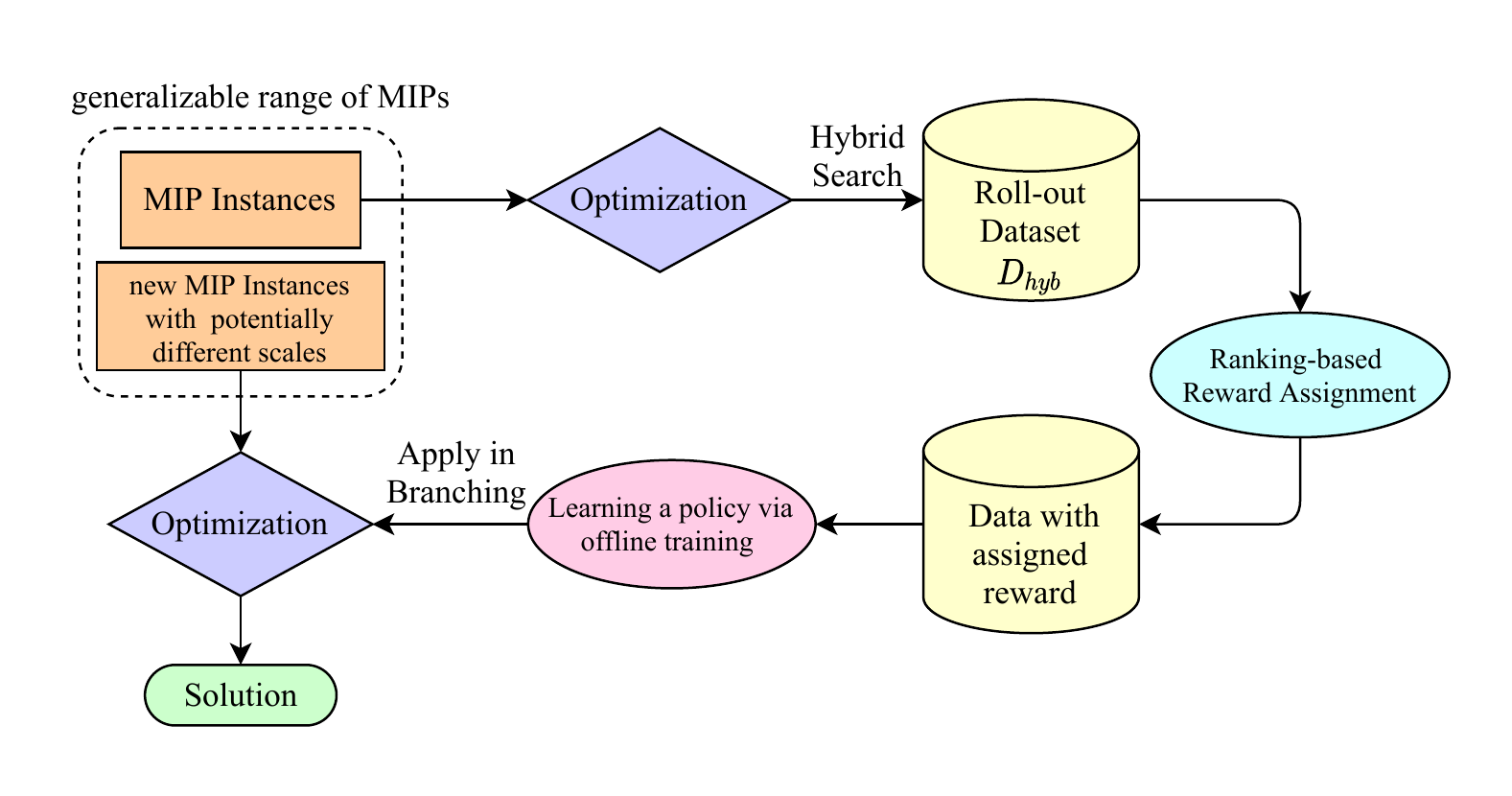}
     \caption{The flowchart of \textsc{Branch Ranking} method.}
     \label{fig-flowchart}
 \end{figure}

We then train our policy $\pi_{\theta}$ via Eq.~(\ref{obj}), and evaluate $\pi_{\theta}$ on the new problem instances. The flowchart of our proposed method is shown in Figure \ref{fig-flowchart}. 

\section{Experiments}
We conduct extensive experiments on the MIP benchmarks, and evaluate our policy compared to other state-of-the-art variable selection heuristics as well as learning-based branching models. By comparative analysis on the experimental results, we try to answer the following research questions:
\begin{itemize}[leftmargin=30pt]
    \item[\textbf{RQ1}:] Is our policy more efficient and robust compared to typical heuristics, and learning-based models which imitate the strong branching strategy? 
    \item[\textbf{RQ2}:] Can our policy generalize to different scales of problem instances of the same MIP classes?
    \item[\textbf{RQ3}:] How does different proportions of long-term and short-term promising samples influence the final performance of our trained polices? 
    \item[\textbf{RQ4}:] Can our policy be deployed in challenging real-world tasks and improve the solution quality given a solving time limit? 
\end{itemize}

\subsection{Experimental Setup}
\subsubsection{Benchmarks used for training and evaluating}
The synthetic MIP benchmarks consist of four classes of NP-hard MIP problems as Gasse et al.~\cite{gasse2019exact}: \emph{set covering}, \emph{combinatorial auction}, \emph{capacitated facility location} and \emph{maximum independent set}, which cover a wide range of real-world decision problems. For each class of MIPs, we randomly generate three scales of problem instances for training and evaluating. Since larger MIPs are usually more difficult to solve, the generated instances are referred to as easy, medium and hard instances according to the problem scales. For a more detailed description of the MIP problems, one can refer to \cite{gasse2019exact}. For each MIP class, we train the learning-based policy only on easy instances, and evaluate on other new easy instances as well as medium and hard instances. For each difficulty, the number of instances generated for evaluation is 20. 


Note that we are also aware of other MIP benchmarks such as MIPLIB or the problem benchmarks of ML4CO, which we do not include in the experiments since our benchmarks are representative of the types of the MIP problems encountered in practice. Moreover, the problem benchmarks of ML4CO or MIPLIB are also based on typical binary MIP problems which shares the similar formulation as our benchmarks. 

We use the open-source MIP solver SCIP 6.0.1 \cite{GleixnerEtal2018ZR} for solving the MIPs and implementing our variable selection policy. The main algorithmic framework for optimizing the MIPs is branch-and-bound, combined with the cutting planes enabled at the root node and some other MIP heuristics. The maximum time limit of solving a problem instance is set to 3600 seconds.



\subsubsection{Metrics}
We consider both the number of processed B\&B nodes and the MIP solution time as the metrics to demonstrate the comprehensive performance of different variable selection policies. For each evaluated policy, we show the average value and the standard deviation of the evaluation results over a number of test instances. Note that the number of processed B\&B nodes are reported only on solved instances (integrality gap is zero) within the time limit.


\subsubsection{Baselines}
We compare \textsc{Branch Ranking} with six baselines.
\begin{itemize}[leftmargin=10pt]
	\item \textbf{Branching heuristics}: we compare against four commonly-used and effective branching heuristics in MIP solvers:
	      \emph{inference} (\textsc{Inference}), \emph{mostinf} (\textsc{Mostinf}), \emph{pseudocost branching} (\textsc{Pscost}) and \emph{reliability pseudocost branching} (\textsc{Relpcost}). One can find a detailed description of the heuristics in the documentation of \cite{GleixnerEtal2018ZR}. 
	\item \textbf{Khalil} \cite{khalil2016learning}: the state representation of MIP is based on available MIP statistics. A SVMrank \cite{joachims2002optimizing} model is used to learn an approximation to strong branching. 
	\item \textbf{GCN} \cite{gasse2019exact}: the state representation of MIP is the bipartite graph, and the policy architecture is based on GCN which takes the MIP state as the inputs. The policy is trained to mimic the strong branching strategy via imitation learning.
\end{itemize}


Note that the method of \cite{sun2020improving} or \cite{etheve2020reinforcement} is not included as a learning-based baseline since the proposed algorithmic framework is on-policy, while in this work, we formulate learning to branch in the offline settings, and propose to derive a branching policy via offline learning, which we consider to be more suitable for the nature of the variable selection problem in B\&B.

\subsubsection{Hyper-parameters}
The architecture of our policy network is based on GCN as \cite{gasse2019exact}. For other hyper-parameters, we set the exploring times $k$ to 30, the ranking parameter $p$ to 10. We fine-tune the data proportion parameter $h$ for each MIP class, and set $h$ to 0.7, 0.9, 0.95, 0.9, 0.9 for \emph{set covering}, \emph{combinatorial auction}, \emph{capacitated facility location} and \emph{maximum independent set}, respectively. For each learning-based branching model, the number of training and validation samples are $50000$ and $5000$, respectively.

\begin{table}
	\centering  
	\caption{Evaluation results of different heuristics and learning-based policies in terms of the solution time and the number of processed B\&B nodes. For each MIP class, the machine learning models are trained on easy instances only. The best evaluation result is highlighted. Among the compared methods, the learning-based ones are marked with $*$.}
	\label{tab:exp_1} 
	\vspace{-5pt}
    \captionsetup{labelformat=empty}
	\caption{Set Covering}
	\resizebox{1.0\textwidth}{!}{
	\begin{tabular}{ccccccc} 
			\toprule  
			\multirow{2}{*}{Method}&  
			\multicolumn{2}{c}{Easy}&\multicolumn{2}{c}{ Medium}&\multicolumn{2}{c}{Hard}\cr  
			\cmidrule(lr){2-3} \cmidrule(lr){4-5} \cmidrule(lr){6-7} 
			& Time & Nodes & Time & Nodes & Time & Nodes\cr  
			\midrule  
            \textsc{Inference} & 16.15$\pm$17.17 &1366$\pm$1636  &689.85$\pm$893.48  & 62956$\pm$91525  &3600.00$\pm$0.00  & N/A\cr  
            \textsc{Mostinf} & 31.22$\pm$33.59 &2397$\pm$2565  &1216.57$\pm$1042.13  & 73955$\pm$91098  &3600.00$\pm$0.00  & N/A\cr  
            
			
            \textsc{Pscost} & 6.96$\pm$2.38 &209$\pm$134  &139.28$\pm$178.43  & 13101$\pm$17965  &2867.07$\pm$882.83  & 128027$\pm$53366\cr  
            
			\textsc{Relpscost}&9.84$\pm$3.07 &\textbf{105$\pm$125}  & 106.21$\pm$112.43 &7223$\pm$9853&2530.76$\pm$1121.30 & 100952$\pm$65534 \cr  
            
	        \textsc{Khalil *} & 8.18$\pm$3.76 &118$\pm$88  &146.58$\pm$185.44  & 6896$\pm$9330  &2866.54$\pm$973.77  & 73780$\pm$33070\cr  

			\textsc{GCN *}&6.99$\pm$1.80 &168$\pm$138& 86.76$\pm$98.56& 5067$\pm$6512&2356.36$\pm$1189.37& 71576$\pm$47806 \cr  
			\textsc{BR (Ours) *} & \textbf{6.93$\pm$1.67} & 163$\pm$129  &\textbf{80.76$\pm$87.97}   & \textbf{4624$\pm$5912}  &\textbf{2199.49$\pm$1170.82}  & \textbf{67784$\pm$46808}\cr  
			\bottomrule  
	\end{tabular}}
	\vspace{5pt}
	\caption{Combinatorial Auction}
	\vspace{-10pt}
	\resizebox{1.0\textwidth}{!}{
	\begin{tabular}{cccccccccc}
		\toprule  
		\multirow{2}{*}{Method}&  
		\multicolumn{2}{c}{Easy}&\multicolumn{2}{c}{ Medium}&\multicolumn{2}{c}{Hard}\cr  
		\cmidrule(lr){2-3} \cmidrule(lr){4-5} \cmidrule(lr){6-7}  
		& Time & Nodes & Time & Nodes & Time & Nodes\cr  
		\midrule  
        \textsc{Inference} & 2.22$\pm$1.40 &638$\pm$922  &25.09$\pm$13.84  & 6043$\pm$4173  &655.30$\pm$695.36  & 91328$\pm$99142\cr  
        \textsc{Mostinf} & 3.20$\pm$3.36 &938$\pm$1569  &332.19$\pm$269.72  & 76349$\pm$62942  &2786.27$\pm$73.02  & 282830$\pm$42980\cr  
        
        \textsc{Pscost} & \textbf{1.97$\pm$0.98} &374$\pm$432  &22.91$\pm$17.12  & 3843$\pm$3593  &386.61$\pm$351.55  & 39135$\pm$37257\cr  
  
		\textsc{Relpscost}&3.18$\pm$1.58 &\textbf{28$\pm$54}  & 19.53$\pm$7.15 & 987$\pm$843&213.66$\pm$229.91 & 14788$\pm$19115 \cr  
 
        \textsc{Khalil *} & 2.17$\pm$1.16 &123$\pm$129  &24.43$\pm$13.66  & 1154$\pm$782  &612.89$\pm$784.94  & 19291$\pm$25459\cr  

		\textsc{GCN *}&\textbf{1.97$\pm$0.60} &103$\pm$105& 12.74$\pm$6.92& 970$\pm$793&201.78$\pm$260.23& 14529$\pm$20819 \cr  
		
		\textsc{BR (Ours) *} & 1.98$\pm$0.62 & 106$\pm$102  &\textbf{12.12$\pm$5.72}  &  \textbf{902$\pm$627}  &\textbf{198.02$\pm$248.70}  & \textbf{14380$\pm$20057}\cr  
		\bottomrule  
	\end{tabular}}
	\vspace{5pt}
	\caption{Capacitated Facility Location}
	\vspace{-10pt}
	\resizebox{1.0\textwidth}{!}{
		\begin{tabular}{cccccccccc}
		\toprule  
		\multirow{2}{*}{Method}&  
		\multicolumn{2}{c}{Easy}&\multicolumn{2}{c}{ Medium}&\multicolumn{2}{c}{Hard}\cr  
		\cmidrule(lr){2-3} \cmidrule(lr){4-5} \cmidrule(lr){6-7}  
		& Time & Nodes & Time & Nodes & Time & Nodes\cr  
		\midrule  
        \textsc{Inference} & 75.54$\pm$75.50 &254$\pm$455  &356.02$\pm$400.02  & 1040$\pm$1214  &756.98$\pm$408.78  & 547$\pm$384\cr  
        \textsc{Mostinf} & 43.88$\pm$33.26 &128$\pm$194  &280.38$\pm$281.19  & 673$\pm$740  &662.62$\pm$375.54  & 393$\pm$273\cr  
		
        \textsc{Pscost} & 40.80$\pm$31.99 &107$\pm$161  &271.28$\pm$241.35  & 673$\pm$637  &644.65$\pm$400.78  & 433$\pm$360\cr 
        
		\textsc{Relpscost}& 42.29$\pm$29.15 & \textbf{72$\pm$131} & 320.14$\pm$286.33 & \textbf{383$\pm$513}& 703.75$\pm$425.64 & \textbf{233$\pm$239}\cr

	    \textsc{Khalil *} & 41.88$\pm$30.64 &106$\pm$155  &282.97$\pm$342.77  & 681$\pm$913  &664.58$\pm$420.50  & 369$\pm$275\cr

		\textsc{GCN *}& 34.03$\pm$33.05 & 160$\pm$214& 255.35$\pm$259.19 & 584$\pm$647 & 638.07$\pm$410.58  & 479$\pm$444\cr  
		
		\textsc{BR (Ours) *} & \textbf{33.92$\pm$33.90}& 169$\pm$239 & \textbf{249.02$\pm$222.21} & 570$\pm$563 & \textbf{633.59$\pm$406.13} & 459$\pm$399 \cr  
		\bottomrule  
	\end{tabular}}
	\vspace{5pt}
	\caption{Maximum Independent Set}
	\vspace{-10pt}
	\resizebox{1.0\textwidth}{!}{
	\begin{tabular}{cccccccc}
		\toprule  
		\multirow{2}{*}{Method}&  
		\multicolumn{2}{c}{Easy}&\multicolumn{2}{c}{ Medium}&\multicolumn{2}{c}{Hard}\cr  
		\cmidrule(lr){2-3} \cmidrule(lr){4-5} \cmidrule(lr){6-7}  
		& Time & Nodes & Time & Nodes & Time & Nodes\cr  
		\midrule 
        \textsc{Inference} & 32.99$\pm$103.48 &16105$\pm$58614  &1807.13$\pm$750.10  & 116957$\pm$84167  &3600.00$\pm$0.00  & N/A\cr  
        \textsc{Mostinf} & 30.34$\pm$91.58 &10084$\pm$35418  &1547.43$\pm$901.47  & 60128$\pm$52244  &3600.00$\pm$0.00  & N/A\cr  
		
        \textsc{Pscost} & 10.11$\pm$13.88 &1565$\pm$3935  &1042.47$\pm$798.46  & 43256$\pm$38143  &3181.14$\pm$455.73  & 66258$\pm$23654\cr  
        
		\textsc{Relpscost}&8.83$\pm$5.11& \textbf{209$\pm$603}  & 136.78$\pm$113.04& 4911$\pm$5774&2462.67$\pm$1391.50  &53636$\pm$38140 \cr

		\textsc{Khalil *} & 10.50$\pm$16.62 &353$\pm$917  &503.94$\pm$712.89  & 7983$\pm$10915  &3023.02$\pm$798.52  & 60800$\pm$23837\cr

		\textsc{GCN *}&7.55$\pm$10.11& 265$\pm$805& 256.23$\pm$509.38& 10948$\pm$23020&2485.27$\pm$1446.23 &53804$\pm$37253 \cr  
		
		\textsc{BR (Ours) *} & \textbf{7.54$\pm$9.53} &260$\pm$742  &\textbf{102.91$\pm$118.77}  & \textbf{3908$\pm$5613}  &\textbf{2320.20$\pm$1457.44}  & \textbf{50460$\pm$37859}\cr 
		\bottomrule  
	\end{tabular}}
\end{table}  

\subsection{Performance Comparison (RQ1 \& RQ2)}
As shown in Table \ref{tab:exp_1}, for medium and hard instances of each MIP class, our proposed \textsc{Branch Ranking} clearly outperforms all the baselines in terms of the solution time, and also leads to the smallest number of processed B\&B nodes on three classes of the MIP problems, \emph{set covering}, \emph{combinatorial auction} and \emph{maximum independent set}. For \emph{capacitated facility location}, \textsc{Branch Ranking} produces a B\&B tree with fewer nodes compared to other learning-based models and heuristics except \textsc{Relpcost}. However, \textsc{Branch Ranking} reduces the solution time notably.

\begin{wrapfigure}{r}{0.45\textwidth}
  \centering 
    \includegraphics[width=0.43\textwidth]{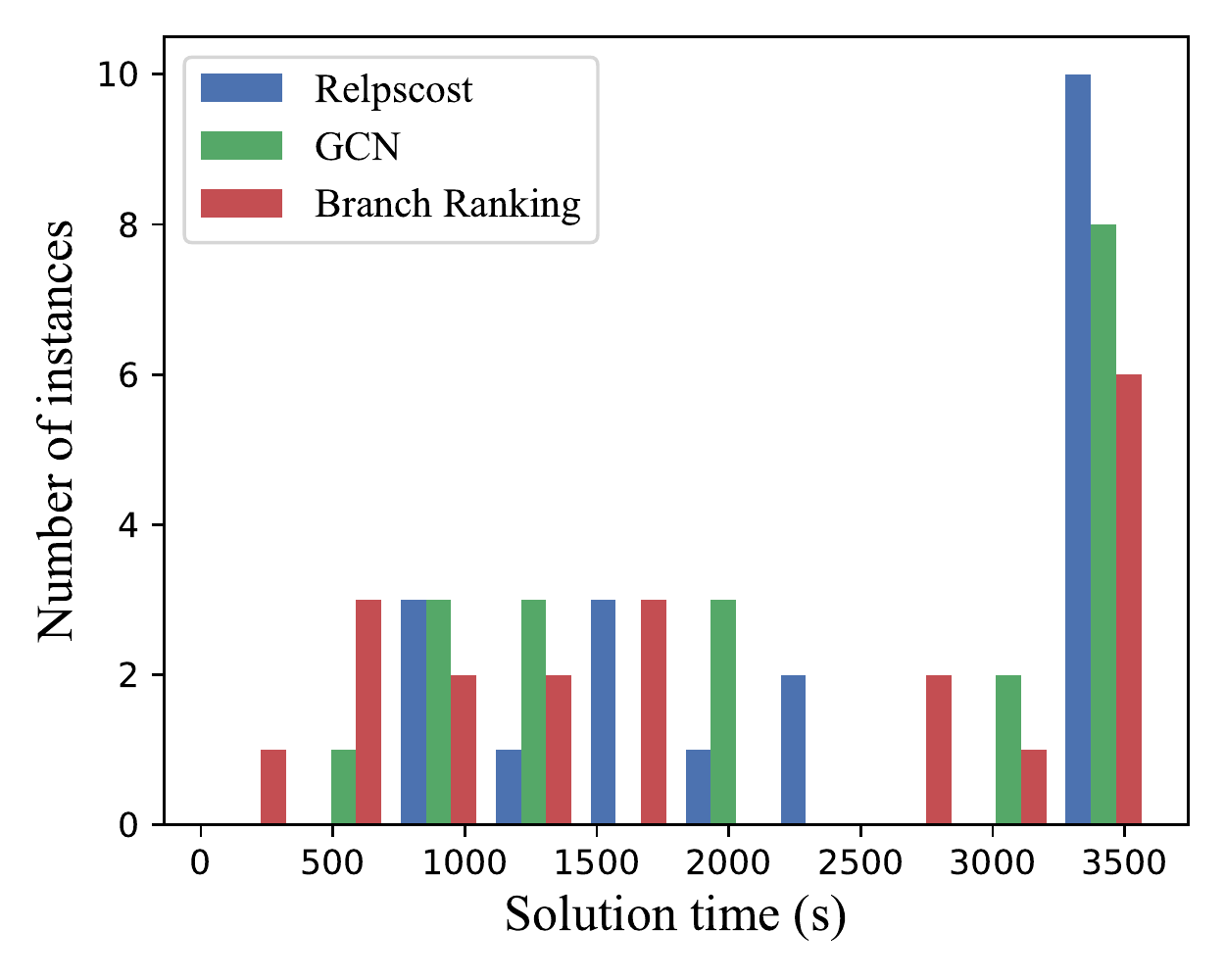}
   \caption{Number of instances with different solution time on the hard set covering problems.}
   \label{fig-hist}
\end{wrapfigure}

As for the easy instances, considering the solution time, \textsc{Branch Ranking} is superior over other baselines on \emph{set covering}, \emph{capacitated facility location} and \emph{maximum independent set}. For the \emph{combinatorial auction} problems, \textsc{Branch Ranking} achieves the second minimum solution time. On account of the number of processed B\&B nodes, \textsc{Branch Ranking} has also shown comparable performance on each MIP class. Note that though \textsc{Relpcost} leads to smaller B\&B trees, it is not competitive regarding the solution time.

The comparative results demonstrate that our proposed policy \textsc{Branch Ranking} is more efficient compared to the state-of-the-art offline learning-based branching models, and also branching heuristics adopted in modern MIP solvers. Moreover, \textsc{Branch Ranking} is also more robust compared to the learning-based models which imitate the strong branching strategy. Specifically, \textsc{Khalil} and \textsc{GCN} suffer from performance fluctuations on the medium and hard \emph{maximum independent set} instances, while \textsc{Branch Ranking} still achieves less solution time and fewer B\&B nodes compared to other well-performed baselines.

Since the learning models are trained only on the easy instances, we are also able to evaluate the generalization ability of different polices. The evaluation results on medium and hard instances show that our proposed \textsc{Branch Ranking} performs remarkably well on larger instances with the same MIP class, and even gains higher performance improvements compared to other baselines when the problem size becomes larger. The results on different scales of instances indicate that \textsc{Branch Ranking} is capable of generalizing over problems with larger scales, and has better generalization ability in comparison to other learning-based branching policies. 

Furthermore, for a better visualization of the generalization results of the well-performed baselines and our proposed \textsc{Branch Ranking}, Figure \ref{fig-hist} shows the number of instances with different solution time for \textsc{Relpcost}, \textsc{GCN} and \textsc{Branch Ranking} on the hard \emph{set covering} problems. The visualization results show that more instances are solved with less solution time using \textsc{Branch Ranking}. Such results are consistent with our previous findings, which also proves that \textsc{Branch Ranking} can better improve the solution process for larger MIP problems.

\subsection{Hyper-parameter Study (RQ3)}
\begin{figure}[!htb]
    \centering
    \begin{subfigure}{\textwidth}
        \centering
            \includegraphics[width=0.6\textwidth]{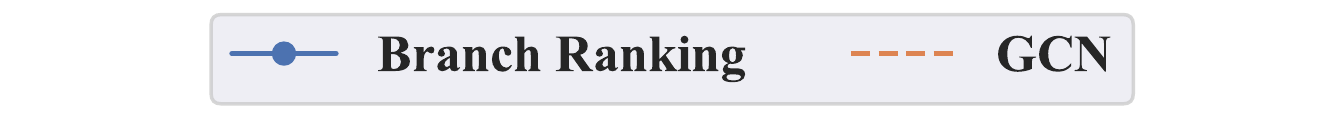} 
    \end{subfigure}
    \\
    \centering
    \begin{subfigure}{\textwidth}
        \centering
            \includegraphics[width=\textwidth]{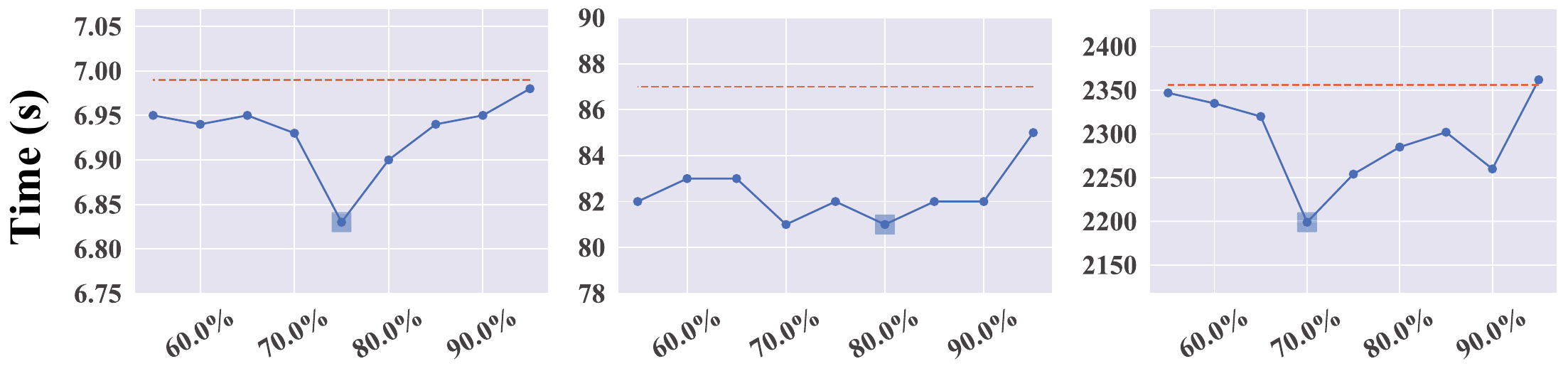} 
    \end{subfigure}
    \\
    \begin{subfigure}{\textwidth}
        \centering
            \includegraphics[width=\textwidth]{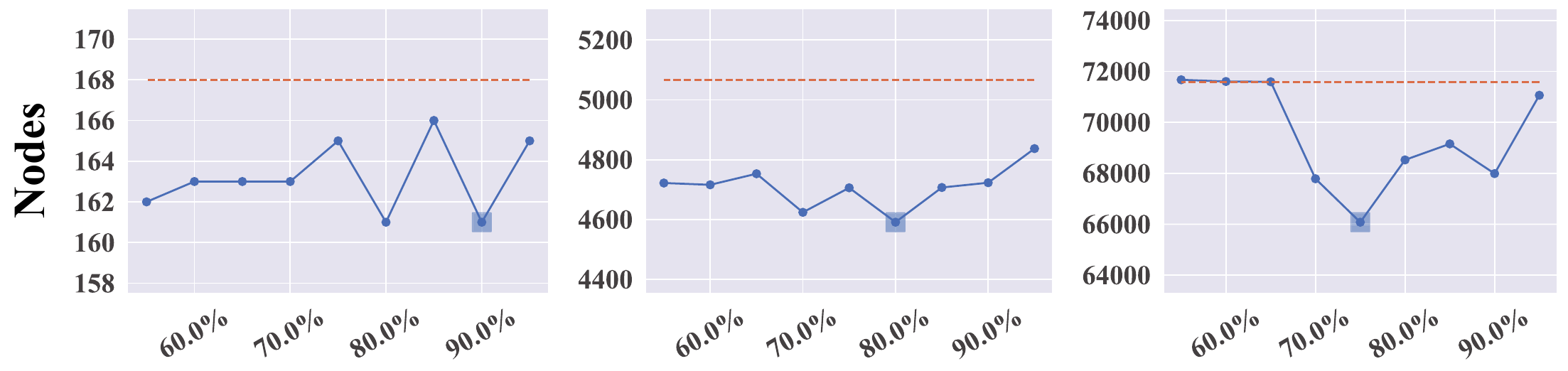} 
    \end{subfigure}
    \\
    \begin{subfigure}{\textwidth}
        \centering
            \includegraphics[width=\textwidth]{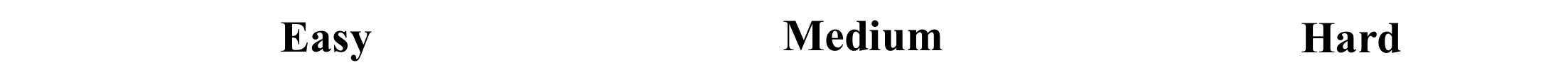} 
    \end{subfigure}
	\caption{Hyper-parameter study results of \textsc{Branch Ranking} with different proportions of short-term promising samples on the set covering problems.}
	\label{fig-ablation-study}
\end{figure}

To better understand the influence of the proportion of long-term or short-term promising samples on the final performance of \textsc{Branch Ranking}, we conduct a hyper-parameter study to evaluate \textsc{Branch Ranking} with different proportions of short-term promising samples on the \emph{set covering} problems. The results are shown in Figure \ref{fig-ablation-study}, in which the dotted line represents the state-of-the-art offline learning-based branching model, \textsc{GCN}, which can be regarded as a policy trained using only the short-term promising samples.

As demonstrated in Figure \ref{fig-ablation-study}, for the easy instances, our proposed \textsc{Branch Ranking} slightly outperforms \textsc{GCN} in terms of solution time and number of processed B\&B nodes under different proportions. Moreover, the policy with an intermediate proportion achieves the best performance considering both the two metrics.

For medium and hard instances, \textsc{Branch Ranking} gains remarkable performance improvements by combining long-term and short-term promising samples. Moreover, the empirical results show that too large or too small short-term sample proportion will worsen the performance of \textsc{Branch Ranking}. Precisely, for the Set Covering problems, we find that setting the short-term sample proportion between 70\% and 80\% often yields a better branching policy.

The results of our hyper-parameter study can be explained from the following aspects. First, as shown by Figure \ref{fig-ablation-study}, the long-term promising samples is beneficial for deriving a more efficient and robust branching policy which can make better decisions from a long-term view. Moreover, using excessive long-term promising samples will probably not produce the best policy, since long-term samples are also more difficult to learn for the policy models. To ease this situation, as introduced in this work, we propose to combine the long-term and short-term promising samples for learning a better variable selection policy for branching.

\subsection{Deployment in Real-World Tasks (RQ4)}
To further verify the effectiveness of our proposed branching policy, we deploy \textsc{Branch Ranking} in the real-world \emph{supply and demand simulation} tasks of Huawei, a global commercial technology enterprise. 
The objective of this task is to find the optimal production planning to meet the current order demand, according to the current raw materials, semi-finished products and the upper limit of production capacity. The basic constraints include the production limit for each production line in each factory, the lot size on product ratio, the order rate, transportation limit, etc.

The collected real-world tasks consist of eight raw cases obtained from the enterprise, each of which can be modeled as a MIP problem instance, and are challenging for the solvers. We regard the raw MIP instances as the test instances, and generate the training instances by adding Gaussian noise to the raw cases. 

In our experiments, we compare \textsc{Branch Ranking} with the default branching strategy incorporated in SCIP (i.e., \textsc{Relpcost} in most cases). We report the MIP solution quality and the MIP solution time on the raw instances. Specifically, the MIP solution quality is measured by the primal-dual gap (lower is better). As for the hyper-parameters, we fine-tune the data proportion parameter $h$ and set it to 0.9, and keep other hyper-parameters the same. The maximum time limit of solving a problem instance is set to 1800 seconds. Note that following the industry procedure, when the problem instance is feasible and not solved to optimal within the time limit, the solver will return an approximate MIP solution.

As shown in Table \ref{tab:exp_r}, since the real-world supply and demand simulation problems are challenging for the solvers, most of the cases are not solved to optimal within the time limit. For case 3 and 4, the MIPs are found infeasible. For case 5, the optimal MIP solution is found for both the baseline and \textsc{Branch Ranking}, while \textsc{Branch Ranking} leads to less solution time. For other cases, though the optimal solution is not found within the time limit, \textsc{Branch Ranking} improves the solution quality notably: the average primal-dual gap reduction ratio has reached 22.74\% compared to the branching strategy incorporated in SCIP. The results further indicate the significance of making good branching decisions. Our proposed \textsc{Branch Ranking} has also shown to be more efficient on difficult real-world tasks.
\addtocounter{table}{-4}
\begin{table}[t]
	\centering  
	\caption{Evaluation results of \textsc{Branch Ranking} and the default branching strategy in SCIP in terms of the solution time and the solution quality (primal-dual gap, the lower is better). Case 3 and 4 are found infeasible.}
	\label{tab:exp_r}
	\vspace{5pt}
	\resizebox{1.0\textwidth}{!}{
	\begin{tabular}{cccccccccccccccccc}
		\toprule  
		\multirow{2}{*}{Method}&  
		\multicolumn{2}{c}{Case 1}&\multicolumn{2}{c}{Case 2}&\multicolumn{2}{c}{Case 3}&\multicolumn{2}{c}{Case 4}&\multicolumn{2}{c}{Case 5}&\multicolumn{2}{c}{Case 6}&\multicolumn{2}{c}{Case 7}&\multicolumn{2}{c}{Case 8}\cr  
		\cmidrule(lr){2-3} \cmidrule(lr){4-5} \cmidrule(lr){6-7} \cmidrule(lr){8-9} \cmidrule{10-11} \cmidrule{12-13} \cmidrule{14-15} \cmidrule{16-17}    
		& Time & Gap & Time & Gap & Time & Gap & Time & Gap & Time & Gap & Time & Gap & Time & Gap & Time & Gap\cr  
		\midrule 
        \textsc{Default} & 1800.0 & 98.59 & 1800.0 & 1.36 & N/A & inf. & N/A & inf. & 460.98 & 0 & 1800.0 & 1.14 & 1800.0 & 0.07 & 1800.0 & 0.21 &\cr
		\textsc{BR (Ours)} & 1800.0 & 62.26 & 1800.0 & 0.87 & N/A & inf. & N/A & inf. & 451.30 & 0 & 1800.0 & 1.14 & 1800.0 & 0.06 & 1800.0 & 0.19  \cr 
		\bottomrule  
	\end{tabular}}
\end{table}

\section{Conclusion}
In this paper, we present an offline RL formulation for variable selection in the branch-and-bound algorithm. To derive a more long-sighted branching policy under such a setting, we propose a top-down hybrid search scheme to collect the offline samples which involves more long-term information. During the policy learning phase, we deploy a ranking-based reward assignment scheme which assigns higher implicit rewards to the long-term or short-term promising samples, and learn a branching policy named \textsc{Branch Ranking} by maximizing the log-likelihood weighted by the assigned rewards. The experimental results on synthetic benchmarks and real-world tasks show that our derived policy \textsc{Branch Ranking} is more efficient and robust compared to the state-of-the-art heuristics and learning-based policies for branching. Furthermore, \textsc{Branch Ranking} can also better generalize to the same class of MIP problems with larger scales.

\section*{Acknowledgements} 
The SJTU team is supported by Shanghai Municipal Science and Technology Major Project (2021SHZDZX0102) and National Natural Science Foundation of China (62076161). The work is also sponsored by Huawei Innovation Research Program.
%
%
%


\begin{thebibliography}{10}
\providecommand{\url}[1]{\texttt{#1}}
\providecommand{\urlprefix}{URL }
\providecommand{\doi}[1]{https://doi.org/#1}

\bibitem{alvarez2017machine}
Alvarez, A.M., Louveaux, Q., Wehenkel, L.: A machine learning-based
  approximation of strong branching. INFORMS Journal on Computing
  \textbf{29}(1),  185--195 (2017)

\bibitem{balcan2018learning}
Balcan, M.F., Dick, T., Sandholm, T., Vitercik, E.: Learning to branch. In:
  International Conference on Machine Learning. pp. 344--353. PMLR (2018)

\bibitem{bengio2021machine}
Bengio, Y., Lodi, A., Prouvost, A.: Machine learning for combinatorial
  optimization: a methodological tour d’horizon. European Journal of
  Operational Research  \textbf{290}(2),  405--421 (2021)

\bibitem{cplex2009v12}
Cplex, I.I.: V12. 1: User’s manual for cplex. International Business Machines
  Corporation  \textbf{46}(53), ~157 (2009)

\bibitem{etheve2020reinforcement}
Etheve, M., Al{\`e}s, Z., Bissuel, C., Juan, O., Kedad-Sidhoum, S.:
  Reinforcement learning for variable selection in a branch and bound
  algorithm. In: International Conference on Integration of Constraint
  Programming, Artificial Intelligence, and Operations Research. pp. 176--185.
  Springer (2020)

\bibitem{fischetti2010heuristics}
Fischetti, M., Lodi, A.: Heuristics in mixed integer programming. Wiley
  Encyclopedia of Operations Research and Management Science  (2010)

\bibitem{gasse2019exact}
Gasse, M., Ch{\'e}telat, D., Ferroni, N., Charlin, L., Lodi, A.: Exact
  combinatorial optimization with graph convolutional neural networks. In:
  Proceedings of the 33rd International Conference on Neural Information
  Processing Systems-Volume 2. pp. 15580--15592 (2019)

\bibitem{glankwamdee2006lookahead}
Glankwamdee, W., Linderoth, J.: Lookahead branching for mixed integer
  programming. Tech. rep., Citeseer (2006)

\bibitem{GleixnerEtal2018ZR}
Gleixner, A., Bastubbe, M., Eifler, L., Gally, T., Gamrath, G., Gottwald, R.L.,
  Hendel, G., Hojny, C., Koch, T., L{\"u}bbecke, M.E., Maher, S.J.,
  Miltenberger, M., M{\"u}ller, B., Pfetsch, M.E., Puchert, C., Rehfeldt, D.,
  Schl{\"o}sser, F., Schubert, C., Serrano, F., Shinano, Y., Viernickel, J.M.,
  Walter, M., Wegscheider, F., Witt, J.T., Witzig, J.: {The SCIP Optimization
  Suite 6.0}. ZIB-Report 18-26, Zuse Institute Berlin (July 2018),
  \url{http://nbn-resolving.de/urn:nbn:de:0297-zib-69361}

\bibitem{gupta2020hybrid}
Gupta, P., Gasse, M., Khalil, E., Mudigonda, P., Lodi, A., Bengio, Y.: Hybrid
  models for learning to branch. Advances in neural information processing
  systems  \textbf{33},  18087--18097 (2020)

\bibitem{joachims2002optimizing}
Joachims, T.: Optimizing search engines using clickthrough data. In:
  Proceedings of the eighth ACM SIGKDD international conference on Knowledge
  discovery and data mining. pp. 133--142 (2002)

\bibitem{khalil2016learning}
Khalil, E.B., Bodic, P.L., Song, L., Nemhauser, G., Dilkina, B.: Learning to
  branch in mixed integer programming. In: Proceedings of the Thirtieth AAAI
  Conference on Artificial Intelligence. pp. 724--731 (2016)

\bibitem{maravs2013routing}
Mara{\v{s}}, V., Lazi{\'c}, J., Davidovi{\'c}, T., Mladenovi{\'c}, N.: Routing
  of barge container ships by mixed-integer programming heuristics. Applied
  Soft Computing  \textbf{13}(8),  3515--3528 (2013)

\bibitem{morrison2016branch}
Morrison, D.R., Jacobson, S.H., Sauppe, J.J., Sewell, E.C.: Branch-and-bound
  algorithms: A survey of recent advances in searching, branching, and pruning.
  Discrete Optimization  \textbf{19},  79--102 (2016)

\bibitem{nair2020solving}
Nair, V., Bartunov, S., Gimeno, F., von Glehn, I., Lichocki, P., Lobov, I.,
  O'Donoghue, B., Sonnerat, N., Tjandraatmadja, C., Wang, P., et~al.: Solving
  mixed integer programs using neural networks. arXiv preprint arXiv:2012.13349
   (2020)

\bibitem{sun2020improving}
Sun, H., Chen, W., Li, H., Song, L.: Improving learning to branch via
  reinforcement learning. In: Learning Meets Combinatorial Algorithms at
  NeurIPS2020 (2020)

\bibitem{zarpellon2021parameterizing}
Zarpellon, G., Jo, J., Lodi, A., Bengio, Y.: Parameterizing branch-and-bound
  search trees to learn branching policies. In: Proceedings of the AAAI
  Conference on Artificial Intelligence. vol.~35, pp. 3931--3939 (2021)

\bibitem{zhu2000minimizing}
Zhu, Z., Heady, R.B.: Minimizing the sum of earliness/tardiness in
  multi-machine scheduling: a mixed integer programming approach. Computers \&
  Industrial Engineering  \textbf{38}(2),  297--305 (2000)

\end{thebibliography}

%

\end{document}